\documentclass[11pt,a4paper]{article}
\usepackage[hyperref]{acl2020}
\usepackage{times}
\usepackage{latexsym}
\usepackage{graphicx}
\usepackage{url}
\usepackage{multirow}
\usepackage{algorithm,algcompatible,amsmath}
\usepackage{mathtools}
\usepackage{amsfonts}
\usepackage{xcolor}
\usepackage{enumitem}
\usepackage{marvosym}
\usepackage{textcomp}
\usepackage{array}
\usepackage{graphicx}
\usepackage{url}
\usepackage{pifont}
\usepackage{amsmath}
\usepackage{booktabs}
\usepackage{makecell}
\usepackage[T1]{fontenc}
\usepackage[utf8]{inputenc}
\usepackage{microtype}

\newcommand{\squishlist}{
\begin{list}{$\bullet$}
{ %\usecounter{Lcount}
\setlength{\itemsep}{0pt}
\setlength{\parsep}{0pt}
\setlength{\topsep}{0pt}
\setlength{\partopsep}{0pt}
\setlength{\leftmargin}{1em}
\setlength{\labelwidth}{1.5em}
\setlength{\labelsep}{0.5em} } }
\newcommand{\squishend}{
\end{list} }

\aclfinalcopy % Uncomment this line for the final submission
\definecolor{my_orange}{RGB}{230, 145, 56}
\definecolor{my_blue}{RGB}{111, 168, 220}
\definecolor{my_red}{RGB}{204, 0, 0}
\definecolor{my_green}{RGB}{106, 168, 79}
\definecolor{my_purple}{RGB}{166, 77, 121}

\title{Reasoning with Latent Structure Refinement\\
for Document-Level Relation Extraction}

\author{Guoshun Nan$^{1}$\thanks{$^{*}$ Equally Contributed.} \ , Zhijiang Guo$^{1}$$^{*}$, Ivan Sekuli\'c$^{2}$\thanks{$^{\dagger}$ Work done during internship at SUTD.}\and Wei Lu$^{1}$\\
  $^{1}$StatNLP Research Group, Singapore University of Technology and Design \\
  $^{2}$Universit\`a della Svizzera italiana \\
  \texttt{guoshun\_nan@sutd.edu.sg, zhijiang\_guo@mymail.sutd.edu.sg} \\
  \texttt{ivan.sekulic@usi.ch, luwei@sutd.edu.sg}\\}

\begin{document}
\maketitle

\begin{abstract}
Document-level relation extraction requires integrating information within and across multiple sentences of a document and capturing complex interactions between inter-sentence entities. However, effective aggregation of relevant information in the document remains a challenging research question. 
Existing approaches construct static document-level graphs based on syntactic trees, co-references or heuristics from the unstructured text to model the dependencies. 
Unlike previous methods that may not be able to capture rich non-local interactions for inference, we propose a novel model that empowers the relational reasoning across sentences by automatically inducing the latent document-level graph. 
We further develop a refinement strategy, which enables the model to incrementally aggregate relevant information for multi-hop reasoning.
Specifically, our model achieves an $F1$ score of 59.05 on a large-scale document-level dataset (DocRED), significantly improving over the previous results, and also yields new state-of-the-art results on the CDR and GDA dataset. Furthermore, extensive analyses show that the model is able to discover more accurate inter-sentence relations. 

% Extensive results on the large-scale document-level (DocRED) dataset show that our model is able to integrate richer local and non-local information based on the refined structure for better reasoning, giving significantly better results than previous approaches.

% Existing approaches perform reasoning based on the static graphs constructed by dependencies, co-references or heuristics.
% Most existing approaches predict relations by considering a local short span of a sequence, which might be limited in providing contextual information for rich inference.
% In this work, we propose a novel model that empowers the multi-hop relational reasoning across sentences by automatically inducing the latent structure, which is iteratively refined based on the interactions between mentions and entities.
% Extensive results on the large-scale DocRED dataset show that our model is able to collect global information based on the refined structure for better inference, giving significantly better results than previous approaches. 
% \maybe{Mention F1 score? Analysis?}
\end{abstract}
\section{Introduction}
\label{sec:1}

Relation extraction aims to detect relations among entities in the text and plays a significant role in a variety of natural language processing applications.
%Relation extraction aims to detect relations among entities in the text. It plays a significant role in a variety of natural language processing applications.
Early research efforts focus on predicting relations between entities within the sentence \citep{Zeng2014RelationCV, Xu2015SemanticRC, Xu2015ClassifyingRV}. However, valuable relational information between entities, such as biomedical findings, is expressed by multiple mentions across sentence boundaries in real-world scenarios \citep{Peng2017CrossSentenceNR}.
Therefore, the scope of extraction in biomedical domain has recently been expanded to cross-sentence level \citep{Quirk2017DistantSF, Gupta2018NeuralRE, song-etal-2019-leveraging}.

% including biomedical knowledge discovery \citep{Quirk2017DistantSF}, knowledge base population \citep{Zhang2017PositionawareAA} and question answering \citep{Yu2017ImprovedNR}. 

% Sentence-level models do not consider interactions across mentions and ignore inter-sentential relations.

\begin{figure}
    \centering
    \includegraphics[scale=0.55]{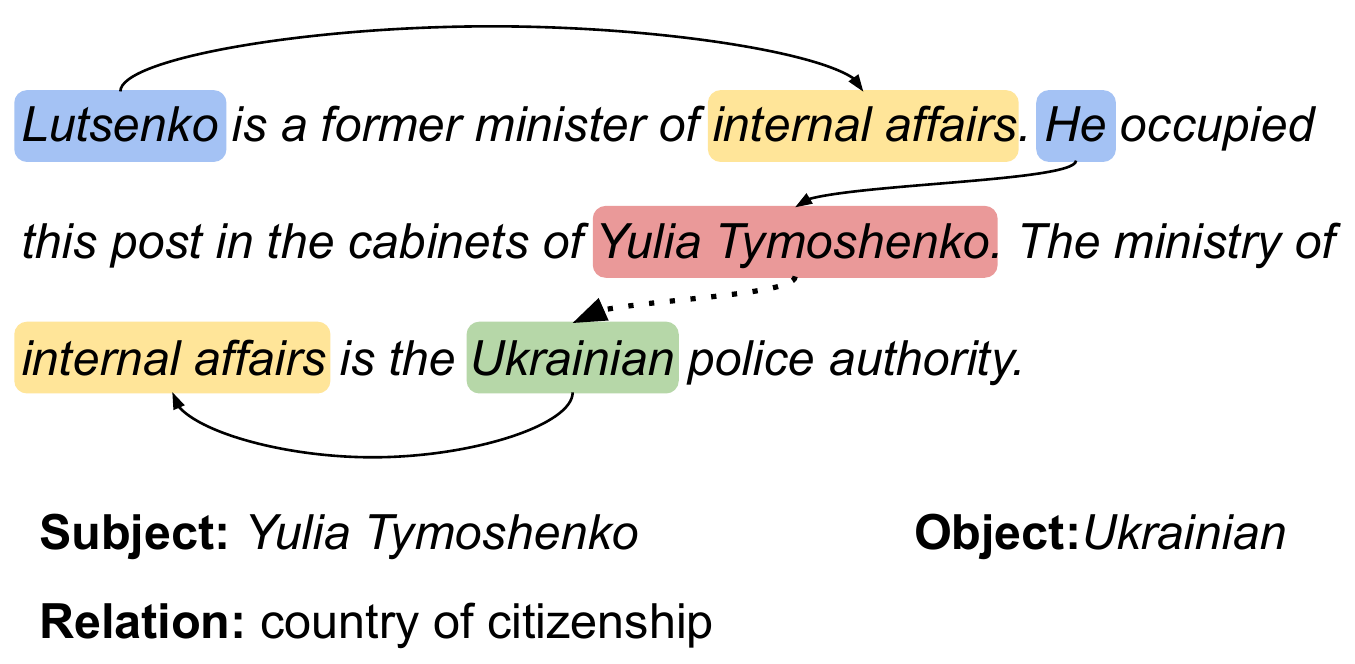}
    \vspace{-2mm}
    \caption{An example adapted from the DocRED dataset. The example has four entities: \textit{Lutsenko}, \textit{internal affairs}, \textit{Yulia Tymoshenko} and \textit{Ukrainian}. Here entity \textit{Lutsenko} has two mentions: \textit{Lutsenko} and \textit{He}. Mentions corresponding to the same entity are highlighted with the same color. The solid and dotted lines represent intra- and inter-sentence relations, respectively.}
    \vspace{-4mm}
    \label{fig:intro}
\end{figure}

A more challenging, yet practical extension, is the document-level relation extraction, where a system needs to comprehend multiple sentences to infer the relations among entities by synthesizing relevant information from the entire document \citep{Jia2019DocumentLevelNR, yao2019DocRED}. Figure \ref{fig:intro} shows an example adapted from the recently proposed document-level dataset DocRED \citep{yao2019DocRED}. In order to infer the inter-sentence relation (i.e., country of citizenship) between \textit{Yulia Tymoshenko} and \textit{Ukrainian}, one first has to identify the fact that \textit{Lutsenko} works with \textit{Yulia Tymoshenko}. Next we identify that \textit{Lutsenko} manages \textit{internal affairs}, which is a \textit{Ukrainian} authority. After incrementally connecting the evidence in the document and performing the step-by-step reasoning, we are able to infer that \textit{Yulia Tymoshenko} is also a \textit{Ukrainian}.

Prior efforts show that interactions between mentions of entities facilitate the reasoning process in the document-level relation extraction. Thus, \citet{Verga2018SimultaneouslyST} and \citet{Jia2019DocumentLevelNR} leverage Multi-Instance Learning \citep{Riedel2010ModelingRA, Surdeanu2012MultiinstanceML}. On the other hand, structural information has been used to perform better reasoning since it models the non-local dependencies that are obscure from the surface form alone. \citet{Peng2017CrossSentenceNR} construct dependency graph to capture interactions among $n$-ary entities for cross-sentence extraction. \citet{Sahu2019IntersentenceRE} extend this approach by using co-reference links to connect dependency trees of sentences to construct the document-level graph. Instead, \citet{christopoulou2019connecting} construct a heterogeneous graph based on a set of heuristics, and then apply an edge-oriented model \citep{christopoulou2018walk} to perform inference. 

% \citet{christopoulou2019connecting} do not rely on any dependency or co-referred structures. They construct a heterogeneous graph based on a set of heuristics then apply an edge-oriented model \citep{christopoulou2018walk} to perform inference. 

Unlike previous methods, where a document-level structure is constructed by co-references and rules, our proposed model treats the graph structure as a latent variable and induces it in an end-to-end fashion. Our model is built based on the structured attention \citep{Kim2017StructuredAN, Liu2017LearningST}. Using a variant of Matrix-Tree Theorem \citep{tutte1984graph, Koo2007StructuredPM}, our model is able to generate task-specific dependency structures for capturing non-local interactions between entities. We further develop an iterative refinement strategy, which enables our model to dynamically build the latent structure based on the last iteration, allowing the model to incrementally capture the complex interactions for better multi-hop reasoning \cite{welbl2018constructing}.

% Combing with the iteratively refine strategy, our model is able to dynamically build the latent structure based on the last iteration, allowing the model to gradually find relevant information for better multi-hop reasoning.

% We further develop an iterative refinement strategy, which
% enables the model to dynamically constructs the latent structure for better information aggregation in the entire document.

% Models making use of structures of the text have proven to be effective in various tasks including relation extraction\citep{miwa-bansal-2016-end, Zhang2018GraphCO}, text summarization \citep{Kim2017StructuredAN, Liu2017LearningST} and question answering\citep{song2018exploring,decao2019naacl}, because structural information models the non-local dependencies that are obscure from the surface form alone. 

Experiments show that our model significantly outperforms the existing approaches on DocRED, a large-scale document-level relation extraction dataset with a large number of entities and relations, and also yields new state-of-the-art results on two popular document-level relation extraction datasets in the biomedical domain. The code and pretrained model are available at \url{https://github.com/nanguoshun/LSR} \footnote{Our model is implemented in PyTorch \citep{Paszke2017AutomaticDI}}.  

Our contributions are summarized as follows:
% \squishlist
\begin{itemize}
\item We construct a document-level graph for inference in an end-to-end fashion without relying on co-references or rules, which may not always yield optimal structures. With the iterative refinement strategy, our model is able to dynamically construct a latent structure for improved information aggregation in the entire document.
\item We perform quantitative and qualitative analyses to compare with the state-of-the-art models in various settings. We demonstrate that our model is capable of discovering more accurate inter-sentence relations by utilizing a multi-hop reasoning module.
\end{itemize}
% \squishend

\section{Model}
\label{sec:3}

\begin{figure*}%[htb]
\centering
\includegraphics[scale=0.47]{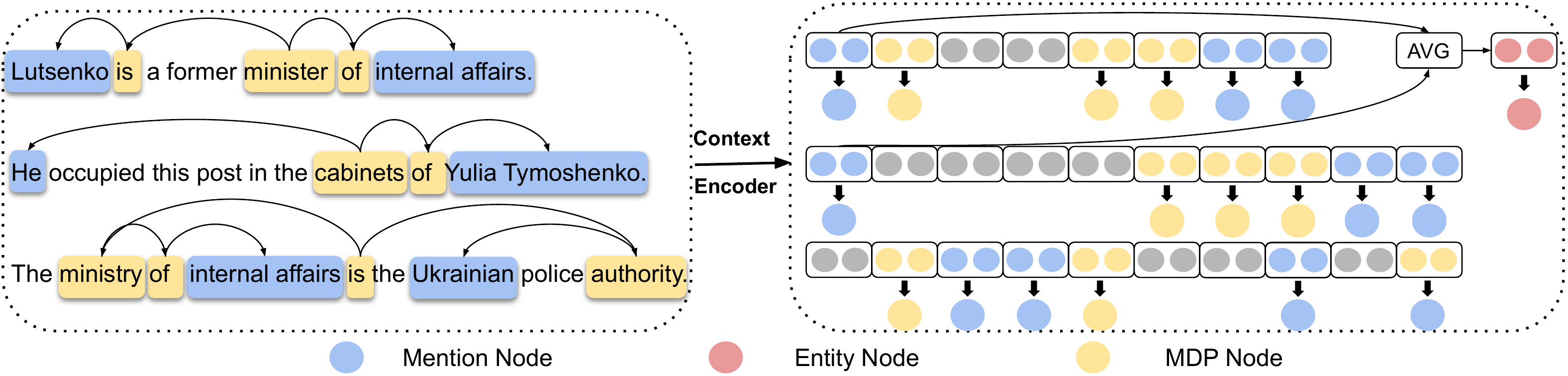}
\vspace{-3mm}
\caption{Overview of the Node Constructor: A context encoder is applied to get the contextualized representations of sentences. The representations of mentions and words in the meta dependency paths are extracted as mention nodes and MDP nodes. An average pooling is used to construct the entity node from the mention nodes. For example, the entity node \textit{Lutsenko} is constructed by averaging representations of its mentions \textit{Lutsenko} and \textit{He}. All figures best viewed in color.} 
% \caption{Overview of the Node Constructor: A context encoder (BiLSTM/BERT) is applied to get the contextualized representations of sentences in the document. The representations of mentions and words on the meta dependency path (MDP) are extracted as mention nodes and MDP nodes. An average pooling is used to construct the entity node from the mention nodes. For example, the entity \textit{Lutsenko} has two mentions \textit{Lutsenko} and \textit{He}. The entity node is constructed by averaging representations of these two mentions.} 
\vspace{-3mm}
\label{fig:constructor}
\end{figure*}

In this section, we present our proposed Latent Structure Refinement (LSR) model for the document-level relation extraction task. Our LSR model consists of three components: node constructor, dynamic reasoner, and classifier. The node constructor first encodes each sentence of an input document and outputs contextual representations. Representations that correspond to mentions and tokens on the shortest dependency path in a sentence are extracted as nodes. The dynamic reasoner is then applied to induce a document-level structure based on the extracted nodes. Representations of nodes are updated based on information propagation on the latent structure, which is iteratively refined. Final representations of nodes are used to calculate classification scores by the classifier.

\subsection{Node Constructor}
\label{ssec:3.1}
Node constructor encodes sentences in a document into contextual representations and constructs representations of mention nodes, entity nodes and meta dependency paths (MDP) nodes, as shown in Figure \ref{fig:constructor}. Here MDP indicates a set of shortest dependency paths for all mentions in a sentence, and tokens in the MDP are extracted as MDP nodes.

\vspace{-1mm}

\subsubsection{Context Encoding}
\label{ssec:3.1.1}
Given a document $\boldsymbol{d}$, each sentence $d_i$ in it is fed to the context encoder, which outputs the contextualized representations of each word in $d_i$. % of the input sentences. 
The context encoder can be a bidirectional LSTM (BiLSTM) \citep{Schuster1997BidirectionalRN} or BERT \citep{Devlin2019BERTPO}. Here we use the BiLSTM as an example: 
\begin{align}
\vspace{-2mm}
    \overleftarrow{\bold{h}_{j}^i} &= \bold{LSTM}_l(\overleftarrow{\bold{h}^i}_{j+1}, \gamma_j^i) \\
    \overrightarrow{\bold{h}_{j}^i} &= \bold{LSTM}_r(\overrightarrow{\bold{h}^i}_{j-1}, \gamma_j^i)
\vspace{-2mm}
\end{align}
where $\overleftarrow{\bold{h}_{j}^i}$, $\overleftarrow{\bold{h}^i}_{j+1}$,   $\overrightarrow{\bold{h}_{j}^i}$ and $\overrightarrow{\bold{h}^i}_{j-1}$ represent the hidden representations of the $j$-th, ($j$+$1$)-th and ($j$-$1$)-th token in the sentence $d_i$ of two directions, and $\gamma_j^i$ denotes the word embedding of the $j$-th token. Contextual representation of each token in the sentence is represented as $\bold{h}_{j}^i = [\overleftarrow{\bold{h}_{j}^i};\overrightarrow{\bold{h}_{j}^i}]$ by concatenating hidden states of two directions, where $\bold{h}_{j}^i \in \mathbb{R}^{d}$ and $d$ is the dimension.

\subsubsection{Node Extraction}
\label{ssec:3.1.2}
We construct three types of nodes for a document-level graph: mention nodes, entity nodes and meta dependency paths (MDP) nodes as shown in Figure \ref{fig:constructor}. Mention nodes correspond to different mentions of entities in each sentence. The representation of an entity node is computed as the average of its mentions. To build a document-level graph, existing approaches use all nodes in the dependency tree of a sentence \citep{Sahu2019IntersentenceRE} or one sentence-level node by averaging all token representations of the sentence \citep{christopoulou2019connecting}. Alternatively, we use tokens on the shortest dependency path between mentions in the sentence. The shortest dependency path has been widely used in the sentence-level relation extraction as it is able to effectively make use of relevant information while ignoring irrelevant information \citep{Bunescu2005ASP, Xu2015SemanticRC, Xu2015ClassifyingRV}. Unlike sentence-level extraction, where each sentence only has two entities, each sentence here may involve multiple mentions. %Formally, let $\bold{u}_i$ denote the contextual representation of the $i$-th node, where $\bold{u}_i \in \mathbb{R}^{d}$.

\begin{figure}[!htb]
\centering
\includegraphics[scale=0.52]{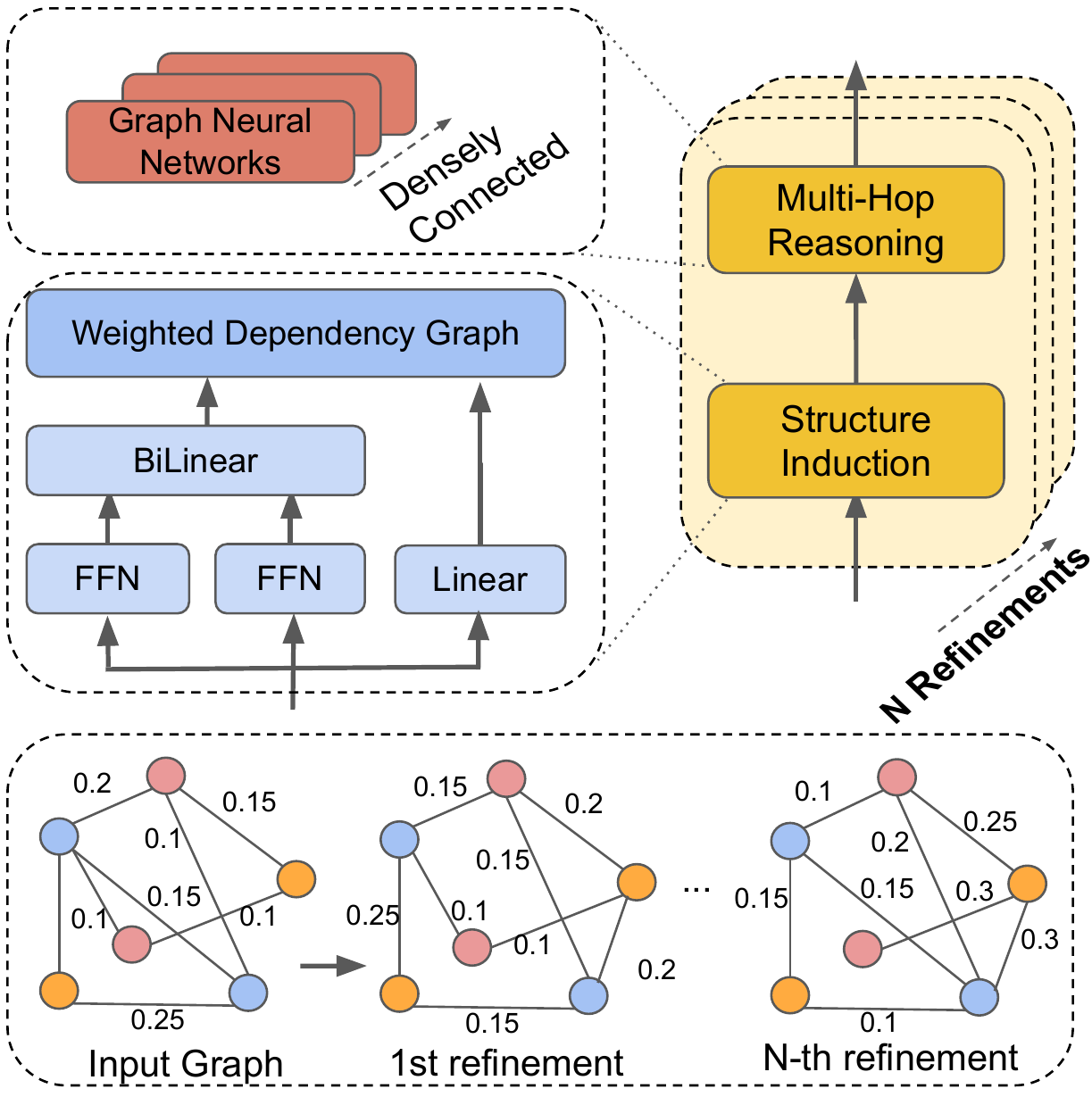}
\vspace{-1mm}
\caption{Overview of the Dynamic Reasoner. Each block consists of two sub-modules: structure induction and multi-hop reasoning. The first module takes the nodes constructed by the Node Constructor as inputs. Representations of nodes are fed into two feed-forward networks before the bilinear transformation. The latent document-level structure is computed by the Matrix-Tree Theorem. The second module takes the structure as input and updates representations of nodes by using the densely connected graph convolutional networks. We stack $N$ blocks which correspond to $N$ times of refinement. Each iteration outputs the latent structure for inference.}
\vspace{-4mm}
\label{fig:reasoner}
\end{figure}

\subsection{Dynamic Reasoner}
\label{ssec:3.2}
The dynamic reasoner has two modules, structure induction and multi-hop reasoning as shown in Figure \ref{fig:reasoner}. The structure induction module is used to learn a latent structure of a document-level graph. The multi-hop reasoning module is used to perform inference on the induced latent structure, where representations of each node will be updated based on the information aggregation scheme. We stack $N$ blocks in order to iteratively refine the latent document-level graph for better reasoning.

\subsubsection{Structure Induction}
\label{ssec:3.2.1}
Unlike existing models that use co-reference links \citep{Sahu2019IntersentenceRE} or heuristics \citep{christopoulou2019connecting} to construct a document-level graph for reasoning, our model treats the graph as a latent variable and induces it in an end-to-end fashion. The structure induction module is built based on the structured attention \citep{Kim2017StructuredAN, Liu2017LearningST}. Inspired by \citet{Liu2017LearningST}, we use a variant of Kirchhoff's Matrix-Tree Theorem \citep{tutte1984graph, Koo2007StructuredPM} to induce the latent dependency structure. 

Let $\bold{u}_i$ denote the contextual representation of the $i$-th node, where $\bold{u}_i \in \mathbb{R}^{d}$, we first calculate the pair-wise unnormalized attention score $\bold{s}_{ij}$ between the $i$-th and the $j$-th node with the node representations $\bold{u}_i$ and $\bold{u}_j$. The score $\bold{s}_{ij}$ is calculated by two feed-forward neural networks and a bilinear transformation:
\begin{align}
    \bold{s}_{ij} = (\tanh(\bold{W}_p \bold{u}_i))^T \bold{W}_b  (\tanh(\bold{W}_c \bold{u}_j))
    \label{eq:dep_ij}
\end{align}
where $\bold{W}_p$ $\in$ $\mathbb{R}^{d \times d}$ and $\bold{W}_c$ $\in$ $\mathbb{R}^{d \times d}$ are weights for two feed-forward neural networks, $d$ is the dimension of the node representations, and $\tanh$ is applied as the activation function.  $\bold{W}_b$ $\in$ $\mathbb{R}^{d \times d}$ are the weights for the bilinear transformation. Next we compute the root score $\bold{s}^r_{i}$ which represents the unnormalized probability of the $i$-th node to be selected as the root node of the structure: 
\begin{align}
    \bold{s}^r_i = \bold{W}_r \bold{u}_i
    \label{eq:dep_root}
\end{align}
where $\bold{W}_r$ $\in$ $\mathbb{R}^{1 \times d}$ is the weight for the linear transformation. Following \citet{Koo2007StructuredPM}, we calculate the marginal probability of each dependency edge of the document-level graph. For a graph $\bold{G}$ with $n$ nodes, we first assign non-negative weights $\bold{P}$ $\in$ $\mathbb{R}^{n \times n}$ to the edges of the graph: %in Equation (\ref{eq: lap1}).

\begin{align}
\vspace{-2mm}
\bold{P}_{ij}=
\begin{cases}
0& \text{if} \ i = j\\
\exp{(\bold{s}_{ij})}& \text{otherwise}
\end{cases}
\label{eq: lap1}
\end{align}
where $\bold{P}_{ij}$ is the weight of the edge between the $i$-th  and the $j$-th node. We then define the Laplacian matrix $\bold{L} \in \mathbb{R}^{n\times n}$ of $\bold{G}$ in Equation (\ref{eq:lap2}), and its variant $\bold{\hat{L}} \in \mathbb{R}^{n\times n}$ in Equation (\ref{eq:lap3}) for further computations \cite{Koo2007StructuredPM}.%reduce the computation complexity of the partition function. %It replaces the first row of $\bold{L} \in \mathbb{R}^{n\times n}$ by the non-negative weights of root nodes.  
\begin{align}
\bold{L}_{ij}=
\begin{cases}
\sum^n_{i'=1} \bold{P}_{i'j} & \text{if} \ i = j\\
-\bold{P}_{ij}& \text{otherwise}
\end{cases}
\label{eq:lap2}
\end{align}

\begin{align}
\bold{\hat{L}}_{ij}=
\begin{cases}
\exp(\bold{s}^r_i) & \text{if} \ i = 1\\
\bold{L}_{ij} & \text{if} \ i > 1
\end{cases}
\label{eq:lap3}
\end{align}

%The marginal probability of dependency edges can be calculated by the weighted sum of dependency trees and the partition function in a differentiable way \cite{Koo2007StructuredPM}. 
We use $\bold{A}_{ij}$ to denote the marginal probability of the dependency edge between the $i$-th and the $j$-th node. Then, $\bold{A}_{ij}$ can be derived based on Equation (\ref{eq:att_ij}), where $\delta$ is the Kronecker delta \cite{Koo2007StructuredPM}.% which is defined in Equation (\ref{eq:delta}).

\begin{align}
\begin{split}
\bold{A}_{ij} = (1 - \delta_{1,j}) \bold{P}_{ij} [\bold{\hat{L}}^{-1}]_{ij} \\
- (1 - \delta_{i,1}) \bold{P}_{ij}[\bold{\hat{L}}^{-1}]_{ji} 
\end{split}
\label{eq:att_ij}
\end{align}
\iffalse
\begin{align}
\delta_{ij}=
\begin{cases}
1 & \text{if} \ i = j\\
0 & \text{otherwise}
\end{cases}
\label{eq:delta}
\end{align}
\fi
\iffalse
where $\bold{L} \in \mathbb{R}^{n\times n}$ is the Laplacian matrix for graph $\bold{G} $ and $\bold{\hat{L}}$ $\in$ $\mathbb{R}^{n \times n}$ is a variant of $\bold{L}$ that takes the root node into consideration, and $\delta$ is the Kronecker delta. 
$\bold{A}_{ij}$ is the marginal probability of the dependency edge between the $i$-th and $j$-th words. 
\fi
Here, $\bold{A} \in \mathbb{R}^{n \times n}$ can be interpreted as a weighted adjacency matrix of the document-level entity graph. 
Finally, we can feed $\bold{A} \in \mathbb{R}^{n \times n}$ into the multi-hop reasoning module to update the representations of nodes in the latent structure.

\subsubsection{Multi-hop Reasoning}
\label{ssec:3.2.2}
Graph neural networks have been widely used in different tasks to perform multi-hop reasoning \citep{song2018exploring, Yang2019AligningCE, Tu2019MultihopRC, kagnet-emnlp19}, as they are able to effectively collect relevant evidence based on an information aggregation scheme. %Our model also leverages GNNs to perform reasoning. 
Specifically, our model is based on graph convolutional networks (GCNs) \citep{Kipf2016SemiSupervisedCW} to perform reasoning.

Formally, given a graph $\bold{G}$ with $n$ nodes, which can be represented with an $n \times n$ adjacency matrix $\bold{A}$ induced by the previous structure induction module, the convolution computation for the node $i$ at the $l$-th layer, which takes the representation $\bold{u}_i^{l-1}$ from previous layer as input and outputs the updated representations $\bold{u}_i^l$, can be defined as:
\begin{align}
    \bold{u}_i^l = \sigma(\sum_{j=1}^{n} \bold{A}_{ij} \bold{W}^l \bold{u}_j^{l-1} + \bold{b}^l)
    \label{eq:gcn}
\end{align}
where $\bold{W}^l$ and $\bold{b}^l$ are the weight matrix and bias vector for the $l$-th layer, respectively. $\sigma$ is the ReLU \citep{nair2010rectified} activation function. $\bold{u}_i^{0} \in \mathbb{R}^{d}$ is the initial contextual representation of the $i$-th node constructed by the node constructor.

Following \citet{dcgcnforgraph2seq19guo}, we use dense connections to the GCNs in order to capture more structural information on a large document-level graph. With the help of dense connections, we are able to train a deeper model, allowing richer local and non-local information to be captured for learning a better graph representation. The computations on each graph convolution layer is similar to Equation (\ref{eq:gcn}).
\iffalse
Mathematically, we first define $\bold{g}_i^l$ as the concatenation of the initial node representations and the node representations produced in layers $1, ..., l-1$:
\begin{align}
    \bold{g}_i^l = [\bold{g}_i^0; \bold{g}_i^1; ... , \bold{g}_i^{l-1}]
\end{align}
The computations on each graph convolution layer are performed based on Equation (\ref{eq:gcn}).
\fi
\iffalse
\begin{align}
    \bold{u}_i^l = \sigma(\sum_{j=1}^{n} \bold{A}_{ij} \bold{W}^l \bold{g}_i^{l-1} + \bold{b}^l)
    \label{eq:dcgcn}
\end{align}
\fi
% where $\bold{W}^l$ is the weight matrix, $\bold{b}^l$ is the bias vector and $\sigma$ is an activation function (e.g. ReLU\citep{nair2010rectified}).

\subsubsection{Iterative Refinement}
\label{ssec:3.2.3}
%\alert{Though structured attention, \citep{Kim2017StructuredAN, Liu2017LearningST} are able to automatically induce a latent structure. However, recent research efforts show that the induced structure is relatively shallow and may not be able to model the complex dependencies for document-level input \citep{Liu2019SingleDS, ferracane2019evaluating}.}
Though structured attention \citep{Kim2017StructuredAN, Liu2017LearningST} is able to automatically induce a latent structure, recent research efforts show that the induced structure is relatively shallow and may not be able to model the complex dependencies for document-level input \citep{Liu2019SingleDS, ferracane2019evaluating}. 
Unlike previous work \citep{Liu2017LearningST} that only induces the latent structure once, we repeatedly refine the document-level graph based on the updated representations, allowing the model to infer a more informative structure that goes beyond simple parent-child relations.

As shown in Figure \ref{fig:reasoner}, we stack $N$ blocks of the dynamic reasoner in order to induce the document-level structure $N$ times. Intuitively, the reasoner induces a shallow structure at early iterations since the information propagates mostly between neighboring nodes. As the structure gets more refined by interactions with richer non-local information, the induction module is able to generate a more informative structure.

\subsection{Classifier}
% After $N$ times of refinement, we obtain hidden representations of all models.
After $N$ times of refinement, we obtain representations of all the nodes. %Given these representations, the goal of is to predict the relation among entities. The representation of an entity is computed as the average of the representations of its mentions. 
Following \citet{yao2019DocRED}, for each entity pair $(\bold{e}_i, \bold{e}_j)$, we use a bilinear function to compute the probability for each relation type $r$ as:
\begin{align}
    P(r|\bold{e}_i, \bold{e}_j) = \sigma (\bold{e}^T_i \bold{W_e} \bold{e}_j + \bold{b}_e)_r
    \label{eq:classifify_bili}
\end{align}
where $\bold{W}_e \in \mathbb{R}^{d \times k \times d}$ and $\bold{b}_e \in \mathbb{R}^{k}$ are trainable weights and bias, with $k$ being the number of relation categories, $\sigma$ is the \textit{sigmoid} function, and the subscript $r$ in the right side of the equation refers to the relation type.  
\section{Experiments}
\label{sec:4}

\subsection{Data}
\label{ssec:4.1}

We evaluate our model on DocRED \citep{yao2019DocRED}, the largest human-annotated dataset for document-level relation extraction, and another two popular document-level relation extraction datasets in the biomedical domain, including Chemical-Disease Reactions (CDR)  \cite{Li2016BioCreativeVC} and Gene-Disease Associations (GDA) \cite{wu2019renet}. DocRED contains $3,053$ documents for training, $1,000$ for development and $1,000$ for test, totally with $132,375$ entities and $56,354$ relational facts. CDR consists of $500$ training instances, $500$ development instances, and $500$ testing instances. GDA contains $29,192$ documents for training and $1,000$ for test. We follow \cite{christopoulou2019connecting} to split training set of GDA into an 80/20 split for training and development.

With more than $40\%$ of the relational facts requiring reading and reasoning over multiple sentences, DocRED significantly differs from previous sentence-level datasets \citep{Doddington2004TheAC, Hendrickx2009SemEval2010T8, Zhang2018GraphCO}. Unlike existing document-level datasets \citep{Li2016BioCreativeVC, Quirk2017DistantSF, Peng2017CrossSentenceNR, Verga2018SimultaneouslyST, Jia2019DocumentLevelNR} that are in the specific biomedical domain considering only the drug-gene-disease relation, DocRED covers a broad range of categories with $96$ relation types.

\begin{table}[h]
\centering
\begin{tabular}{ll}
\toprule
%\textbf{Hyper Parameters} & \textbf{Value}  \\
%\hline
Batch size       & 20      \\
Learning rate    & 0.001        \\
Optimizer        & Adam       \\
Hidden size & 120 \\
Induction block number & 2 \\
GCN dropout & 0.3\\
\bottomrule
\end{tabular}
\caption{\label{tab:hyper} Hyper-parameters of LSR.}
\vspace{-4mm}
\end{table}
\vspace{-2mm}
\subsection{Setup}
\label{ssec:4.2}
We use spaCy\footnote{\url{https://spacy.io/}} to get the meta dependency paths of sentences in a document. Following \citet{yao2019DocRED} and \citet{wang2019fine}, we use the GloVe \citep{Pennington2014GloveGV} embedding with BiLSTM, and Uncased BERT-Base \citep{Devlin2019BERTPO} as the context encoder. All hyper-parameters are tuned based on the development set. We list some of the important hyper-parameters in Table \ref{tab:hyper}. 

Following \citet{yao2019DocRED}, we use $F_1$ and Ign $F_1$ as the evaluation metrics. Ign $F_1$ denotes $F_1$ scores excluding relational facts shared by the training and dev/test sets. $F_1$ scores for intra- and inter-sentence entity pairs are also reported. Evaluation on the test set is done through CodaLab\footnote{\url{https://competitions.codalab.org/competitions/20717}}.

\subsection{Main Results}
\label{ssec:4.3}
\begin{table*}[ht]
% \small
\centering
\scalebox{0.80}{
\begin{tabular}{lcccccc}
\toprule
 & \multicolumn{4}{c}{Dev} & \multicolumn{2}{c}{Test} \\
 \cmidrule(l{5pt}r{5pt}){2-5} \cmidrule(l{5pt}r{5pt}){6-7}
Model & Ign $F1$ & $F1$ & Intra-$F1$ & Inter-$F1$ & Ign $F1$ & $F1$ \\ 
\midrule
CNN \cite{yao2019DocRED}  & 41.58 & 43.45 & 51.87$^*$& 37.58$^*$& 40.33 & 42.26 \\
LSTM \cite{yao2019DocRED} &  48.44 & 50.68& 56.57$^*$& 41.47$^*$ & 47.71 & 50.07 \\
BiLSTM \cite{yao2019DocRED}  & 48.87 & 50.94& 57.05$^*$ & 43.49$^*$ & 48.78 & 51.06 \\
ContexAware \cite{yao2019DocRED} & \textbf{48.94} & 51.09& 56.74$^*$ & 42.26$^*$ & 48.40 & 50.70 \\
\midrule
GCNN $\text{\ding{168}}$ \cite{Sahu2019IntersentenceRE} & 46.22 & 51.52& 57.78\color{white}$^*$ & 44.11\color{white}$^*$ & 49.59 & 51.62 \\
EoG $\text{\ding{168}}$ \cite{christopoulou2019connecting} & 45.94 & 52.15& 58.90\color{white}$^*$ & 44.60\color{white}$^*$ & 49.48 & 51.82 \\
GAT $\text{\ding{168}}$ \cite{velickovic2018graph} & 45.17 & 51.44 & 58.14\color{white}$^*$ & 43.94\color{white}$^*$ & 47.36 & 49.51 \\
AGGCN $\text{\ding{168}}$ \cite{Guo2019AttentionGG}& 46.29 & 52.47 & 58.76\color{white}$^*$ &
45.45\color{white}$^*$ & 48.89 & 51.45 \\
\midrule
GloVe+LSR  & 48.82  & \textbf{55.17}& \textbf{60.83}\color{white}$^*$ & \textbf{48.35}\color{white}$^*$ & \textbf{52.15} & \textbf{54.18} \\
\midrule
BERT \cite{wang2019fine} & - & 54.16& 61.61$^*$ & 47.15$^*$ & - & 53.20 \\
Two-Phase BERT \cite{wang2019fine} & - & 54.42 & 61.80$^*$ & 47.28$^*$ & - & 53.92 \\
BERT+LSR  & \textbf{52.43} & \textbf{59.00}& \textbf{65.26}\color{white}$^*$ & \textbf{52.05}\color{white}$^*$ & \textbf{56.97} & \textbf{59.05} \\
\bottomrule
\end{tabular}}
\vspace{-1mm}
\caption{\label{tbl:main_results} Main results on the development and the test set of DocRED: Models with $\text{\ding{168}}$ are adapted to DocRED based on their open implementations. Results with $*$ are computed based on re-trained models as we need to evaluate $F_1$ for both intra- and inter-sentence setting, which are not given in original papers. }
\vspace{-4mm}
\end{table*}

We compare our proposed LSR with the following three types of competitive models on the DocRED dataset, and show the main results in  Table \ref{tbl:main_results}.

\squishlist
\item \textbf{Sequence-based Models.} These models leverage different neural architectures to encode sentences in the document, including convolutional neural networks (CNN) \cite{Zeng2014RelationCV}, LSTM, bidirectional LSTM (BiLSTM) \cite{cai-etal-2016-bidirectional} and attention-based LSTM (ContextAware) \cite{sorokin2017context}.
\item \textbf{Graph-based Models.} These models construct task-specific graphs for inference. GCNN \citep{Sahu2019IntersentenceRE} constructs a document-level graph by co-reference links, and then applies relational GCNs for reasoning. EoG \citep{christopoulou2019connecting} is the state-of-the-art document-level relation extraction model in biomedical domain. EoG first uses heuristics to construct the graph, then leverages an edge-oriented model to perform inference. GCNN and EoG are based on static structures. GAT \citep{velickovic2018graph} is able to learn the weighted graph structure based on a local attention mechanism. AGGCN \citep{Guo2019AttentionGG} is the state-of-the-art sentence-level relation extraction model, which constructs the latent structure by self-attention. These two models are able to dynamically construct task-specific structures.
\item \textbf{BERT-based Models.} These models fine-tune BERT \citep{Devlin2019BERTPO} for DocRED. Specifically, Two-Phase BERT \cite{wang2019fine} is the best reported model. It is a pipeline model, which predicts if the relation exists between entity pairs in the first phase and predicts the type of the relation in the second phase.
\squishend

As shown in Table \ref{tbl:main_results}, LSR with GloVe achieves 54.18 $F_1$ on the test set, which is
the new state-of-the-art result for models with GloVe.  In particular, our model consistently
outperforms sequence-based models by a significant margin. For example, LSR improves upon the best sequence-based model BiLSTM by $3.1$ points in terms of $F_1$.
This suggests that models which directly encode the entire document are unable to capture the inter-sentence relations present in documents.

Under the same setting, our model consistently outperforms graph-based models based on static graphs or attention mechanisms. Compared with EoG, our LSR model achieves 3.0 and 2.4 higher $F_1$ on development and test set, respectively. We also have similar observations for the GCNN model, which shows that a static document-level graph may not be able to capture the complex interactions in a document. The dynamic latent structure induced by LSR captures richer non-local dependencies. Moreover, LSR also outperforms GAT and AGGCN. This empirically shows that compared to the models that use local attention and self-attention \cite{velickovic2018graph, Guo2019AttentionGG}, LSR can induce more informative document-level structures for better reasoning. Our LSR model also shows its superiority under the setting of Ign $F_1$. 
% These results demonstrate the importance of explicitly model the document-level structure.

In addition, LSR with GloVe obtains better results than two BERT-based models. This empirically shows that our model is able to capture long-range dependencies even without using powerful context encoders. Following \citet{wang2019fine}, we leverage BERT as the context encoder. As shown in Table \ref{tbl:main_results},  our LSR model with BERT achieves a 59.05 $F_1$ score on DocRED, which is a new state-of-the-art result. As of the ACL deadline on the 9th of December 2019, we held the first position on the CodaLab scoreboard under the alias \textit{diskorak}. %\footnote{We held the first place for more than $30$ days prior to the submission deadline of ACL under an alias \textit{diskorak}.}. 
% These results show that our model is more effective in terms of using BERT. 

\subsection{Intra- and inter-sentence performance}
%\input{tables/intra_inter.tex}
%The DocRED requires reasoning across sentences for relation prediction.To comply with the nature of the document-level relation extraction task, 
In this subsection, we analyze intra- and inter-sentence performance on the development set.
An entity pair requires inter-sentence reasoning if the two entities from the same document have no mentions in the same sentence. In DocRED's development set, about $45\%$ of entity pairs require information aggregation over multiple sentences.

%The comparison with the retrained models is also presented in the Table \ref{tbl:main_results}. %The re-implementations and the scores of the retrained models are indicated by $\text{\ding{168}}$ and $*$, respectively.
Under the same setting, our LSR model outperforms all other models in both intra- and inter-sentence setting. 
The differences in $F_1$ scores between LSR and other models in the inter-sentence setting tend to be larger than the differences in the intra-sentence setting.
%While the difference in $F_1$ scores between the baselines and the proposed LSR is around $3$ points in intra-sentence setting, the difference in inter-sentence setting is larger -- around $5$ points of $F_1$. 
These results demonstrate that the majority of LSR's superiority comes from the inter-sentence relational facts, suggesting that the latent structure induced by our model is indeed capable of synthesizing the information across multiple sentences of a document. 

Furthermore, LSR with GloVe also proves better in the inter-sentence setting compared with two BERT-based \cite{wang2019fine} models, indicating latent structure's superiority in resolving long-range dependencies across the whole document compared with the BERT encoder. %Being a strong text encoder capable of tackling multiple reading comprehension tasks \citep{Devlin2019BERTPO}, it is not surprising that the BERT-based model \cite{wang2019fine} outperforms LSR with GloVe in the intra-sentence setting.

\subsection{Results on the Biomedical Datasets}
\begin{table}[]
\centering
% \small
\scalebox{0.8}{
\begin{tabular}{lccc}
\toprule
Model & $F1$ & Intra-$F1$ & Inter-$F1$ \\
\midrule
\citet{Gu2017ChemicalinducedDR}  & 61.3    & 57.2     & 11.7     \\
\citet{Nguyen2018ConvolutionalNN}  & 62.3    & -     & -     \\
\citet{Verga2018SimultaneouslyST}    & 62.1    & -     & -     \\
\midrule
\citet{Sahu2019IntersentenceRE}    & 58.6    & -     & -     \\
\citet{christopoulou2019connecting}      & 63.6    & 68.2  & 50.9  \\
\midrule
LSR    & 61.2    & 66.2  & 50.3 \\
LSR w/o MDP Nodes & \textbf{64.8}    & \textbf{68.9}  & \textbf{53.1} \\
\hline
\hline
\citet{Peng2016ImprovingCD} & 63.1    & -     & -     \\
\citet{Li2016CIDExtractorAC} & 67.3    & 58.9     & -     \\
\citet{panyam2018exploiting}    & 60.3    & 65.1     & 45.7     \\
\citet{Zheng2018AnEN} & 61.5    & -     & -     \\
\bottomrule
\end{tabular}}
\vspace{-1mm}
\caption{\label{tbl:cdr}  Results on the test set of the CDR dataset. The methods below the double line take advantage of additional training data and/or incorporate external tools.}
\vspace{-5mm}
\end{table}

Table \ref{tbl:cdr} depicts the comparisons with state-of-the-art models on the CDR dataset. \citet{Gu2017ChemicalinducedDR, Nguyen2018ConvolutionalNN, Verga2018SimultaneouslyST} leverage sequence-based models. Convolutional neural networks and self-attention networks are used as the encoders. \citet{Sahu2019IntersentenceRE, christopoulou2019connecting} use graph-based models. As shown in Table \ref{tbl:cdr}, our LSR performs worse than the state-of-the-art models. It is challenging for an off-the-shelf parser to get high quality dependency trees in the biomedical domain, as we observe that the MDP nodes extracted by the spaCy parser from the CDR dataset contains much less informative context compared with the nodes from DocRED. Here we introduce a simplified LSR model indicated as ``LSR w/o MDP Nodes'' , which removes the MDP nodes and builds a fully-connected graph using all tokens of a document. It shows that ``LSR w/o MDP Nodes'' consistently outperforms sequence-based and graph-based models, indicating the effectiveness of the latent structure. Moreover, the simplified LSR outperforms most of the models with external resources, except for \citet{Li2016CIDExtractorAC}, which leverages co-training with additional unlabeled training data. We believe such a setting also benefits our LSR model.

%The performance degradation is caused the biomedical domain,
\begin{table}[]
\centering
\small
\scalebox{0.82}{
\begin{tabular}{lccc}
\toprule
Model & $F1$ & Intra-$F1$ & Inter-$F1$ \\
\midrule
NoInf \cite{christopoulou2019connecting}    & 74.6    & 79.1  & 49.3  \\
Full \cite{christopoulou2019connecting}    & 80.8    & 84.1  & \textbf{54.7}  \\
EoG \cite{christopoulou2019connecting}    & 81.5    & 85.2  & 50.0  \\
\midrule
LSR     & 79.6    & 83.1  & 49.6 \\
LSR w/o MDP Nodes & \textbf{82.2}    & \textbf{85.4}  & 51.1 \\
\bottomrule
\end{tabular}}
\vspace{-2mm}
\caption{\label{tbl:gda} Results on the test set of the GDA dataset.}
\vspace{-1mm}
\end{table}

\begin{figure}
    \centering
    \includegraphics[width=200pt]{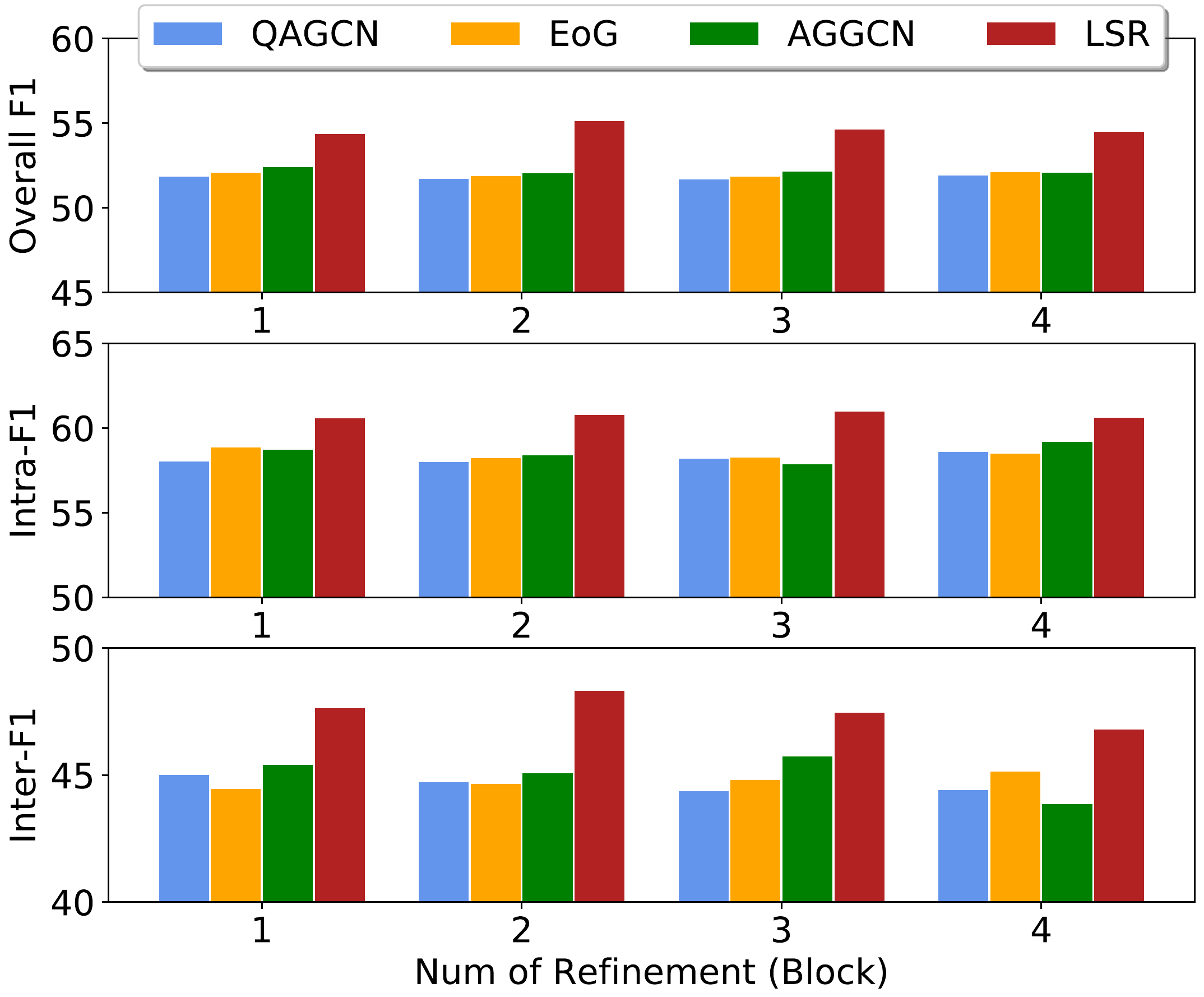}
    \vspace{-1mm}
    \caption{Intra- and inter-sentence $F1$ for different graph structures in QAGCN, EoG, AGGCN and LSR. The number of refinements is ranging from 1 to 4.}
    \label{fig:score}
    \vspace{-6mm}
\end{figure}

 Table \ref{tbl:gda} shows the results on the distantly supervised GDA dataset. Here ``Full'' indicates EoG model with a fully connected graph as the inputs, while ``NoInf'' is a variant of EoG model without inference component \cite{christopoulou2018walk}. The simplified LSR model achieves the new state-of-the-art result on GDA. The ``Full'' model \cite{christopoulou2019connecting} yields a higher $F1$ score on the inter-sentence setting while having a relatively low score on the intra-sentence. It is likely because that this model neglects the differences between relations expressed within the sentence and across sentences. 
%We hypothesize that the neural models are able to better balance the predictions on these two settings by inducing the weighted document-level graph. 

% The comparisons also confirm our hypothesis that the main improvement of LSR comes from inter-sentence relation facts, showing the superior our model in synthesizing cross-sentence contextual information.   

\begin{figure*}
    \centering
    \includegraphics[width=435pt]{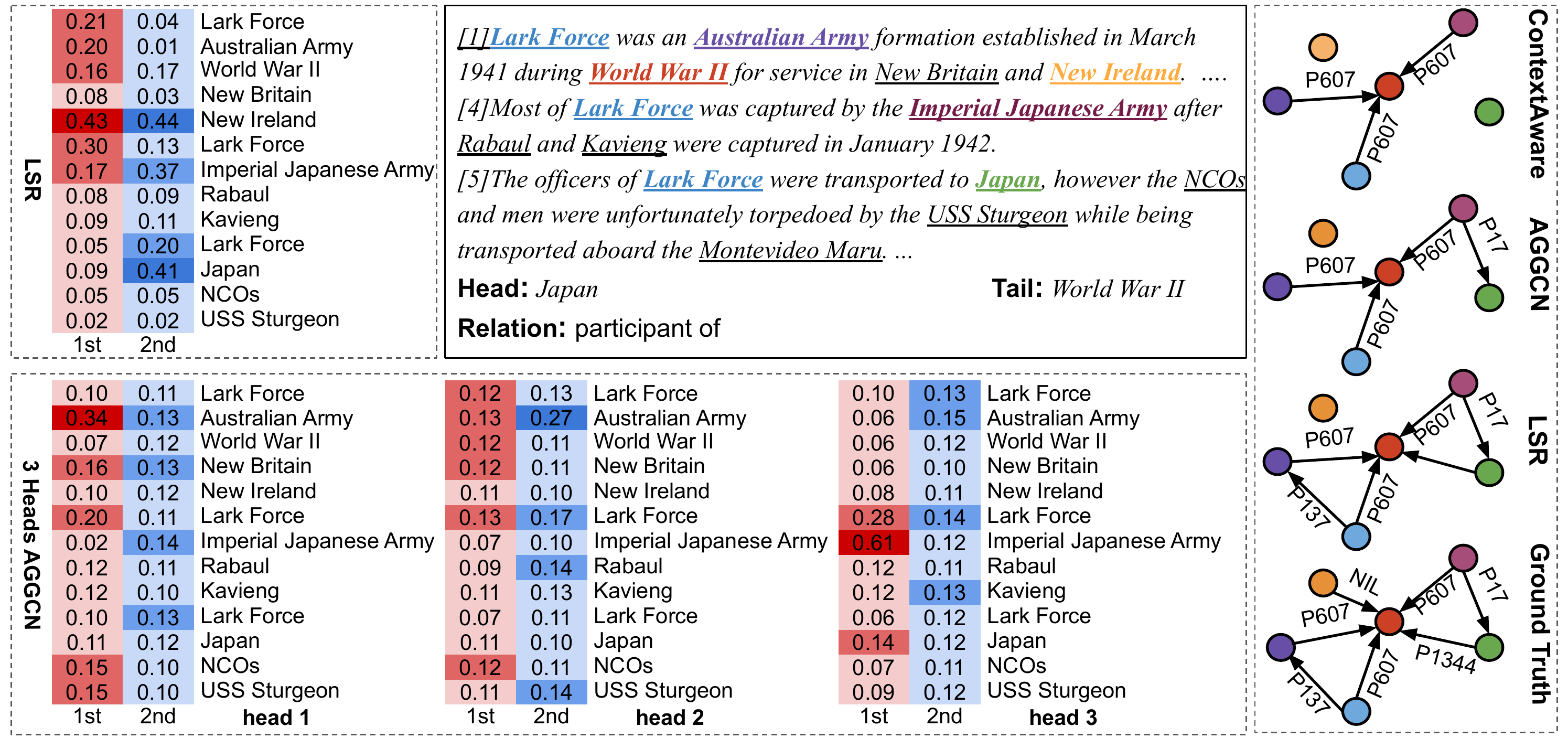}
    \vspace{-1mm}
    \caption{Case study of an example from the development set of DocRED. We visualize the reasoning process for predicting the relation of an entity pair $\langle$\textit{Japan}, \textit{World War II}$\rangle$ by LSR and AGGCN in two refinement steps, using the attention scores of the mention \textit{World War II} in each step. We scale all attention scores by 1000 to illustrate them more clearly. Some sentences are omitted due to space limitation.}
    \label{fig:case}
    \vspace{-4mm}
\end{figure*}

\subsection{Model Analysis}
\label{ssec:4.4}
In this subsection, we use the development set of DocRED to demonstrate the effectiveness of the latent structure and refinements.  
\subsubsection{Does Latent Structure Matter?}
We investigate the extent to which the latent structures, that are induced and iteratively refined by the proposed dynamic reasoner, help to improve the overall performance. We experiment with the three different structures defined below. For fair comparisons, we use the same GCN model to perform multi-hop reasoning for all these structures.

\paragraph{Rule-based Structure:} We use the rule-based structure in EoG \cite{christopoulou2019connecting}. Also, We adapt rules from \citet{de2018question} for multi-hop question answering, i.e., each mention node is connected to its entity node and to the same mention nodes across sentences, while mention nodes and MDP nodes which reside in the same sentence are fully connected. The model is termed QAGCN. 

\paragraph{Attention-based Structure:} This structure is induced by AGGCN \cite{Guo2019AttentionGG} with multi-head attention \citep{Vaswani2017AttentionIA}. We extend the model from sentence-level to document-level.

We explore multiple settings of these models with different block numbers ranging from 1 to 4, where a block is composed of a graph construction component and a densely connected GCN component. As shown in Figure \ref{fig:score}, LSR outperforms QAGCN, EoG and AGGCN in terms of overall $F_1$. This empirically confirms our hypothesis that the latent structure induced by LSR is able to capture a more informative context for the entire document.

\subsubsection{Does Refinement Matter?}
As shown in Figure \ref{fig:score}, our LSR yields the best performance in the second refinement, outperforming the first induction by 0.72\% in terms of overall $F_1$. This indicates that the proposed LSR is able to induce more accurate structures by iterative refinement. However, too many iterations may lead to an $F_1$ drop due to over-fitting. %We leave further investigation as a part of our future work.Compared with intra-F1, the increase of inter-F1 from first refinement to the second refinement is much more significant. This indicates that the inter-sentence entity pairs benefit from our dynamic reasoner more than the intra-sentence pairs.

\begin{table}[]
\small
\scalebox{0.96}{
\begin{tabular}{lccc}
\toprule
Model & $F1$ & Intra-$F1$ & Inter-$F1$ \\
\midrule
Full model & 55.17 & 60.83 & 48.35 \\
- 1 Refinement  & 54.42 & 60.46 & 47.67  \\
- 2 Structure Induction  & 51.91 & 58.08 & 45.04 \\
- 1 Multi-hop Reasoning & 54.49 & 59.75 & 47.49 \\
- 2 Multi-hop Reasoning & 54.24 & 60.58 & 47.15 \\
- MDP nodes & 54.20 & 60.54 & 47.12 \\
\bottomrule
\end{tabular}}
\vspace{-2mm}
\caption{\label{tbl:ablation} Ablation study of LSR on DocRED.}
\vspace{-5mm}
\end{table}
\subsection{Ablation Study}
Table \ref{tbl:ablation} shows $F_1$ scores of the full LSR model and with different components turned off one at a time.
We observe that most of the components contribute to the main model, as the performance deteriorates with any of the components missing. The most significant difference is visible in the structure induction module. Removal of structure induction part leads to a 3.26 drop in terms of $F_1$ score. This result indicates that the latent structure plays a key role in the overall performance.  %Using only 1 step refinement results in 0.75\% drop, indicating that iterative refinement indeed improves the performance. 
%4). Case Study or Qualitative Results.
%4.1 Reasoning Process compared with context-aware LSTM/BERT/our model/Gold. Look at Table 4 in GP-GNNs paper, we should get this one.
%\input{tables/case_studies.tex}
%\input{tables/graphics_case.tex}

\subsection{Case Study}
In Figure \ref{fig:case}, we present a case study to analyze why the latent structure induced by our proposed LSR performs better than the structures learned by AGGCN. We use the entity \textit{World War II} to illustrate the reasoning process and our goal here is to predict the relation of the entity pair $\langle$\textit{Japan}, \textit{World War II}$\rangle$. As shown in Figure \ref{fig:case}, in the first refinement of LSR, \textit{Word War II} interacts with several local mentions with higher attention scores, \text{e.g.,} \textit{0.30} for the mention \textit{Lake Force}, which will be used as a bridge between the mention \textit{Japan} and \textit{World War II}. %In this step, the attention scores show that information is mainly propagated within a sentence. 
In the second refinement, the attention scores of several non-local mentions, such as \textit{Japan} and \textit{Imperial Japanese Army}, significantly increase from \textit{0.09} to \textit{0.41}, and \textit{0.17} to \textit{0.37}, respectively, indicating that information is propagated globally at this step. With such intra- and inter-sentence structures, the relation of the entity pair $\langle$\textit{Japan}, \textit{World War II}$\rangle$ can be predicted as
\textit{``participant of''}, which is denoted by \textit{P1344}. Compared with LSR, the attention scores learned by AGGCN are much more balanced, indicating that the model may not be able to construct an informative structure for inference, e.g., the highest score is \textit{0.27} in the second head, and most of the scores are near \textit{0.11}. 

We also depict the predicted relations of ContextAware, AGGCN and LSR on the graph on the right side of the Figure \ref{fig:case}. Interested reader could refer to \cite{yao2019DocRED} for the definition of a relation, such as \textit{P607}, \textit{P17}, \textit{etc.} The LSR model proves capable of filling out the missing relation for $\langle$\textit{Japan}, \textit{World War II}$\rangle$ that requires reasoning across sentences. %, while other three models fail to predict the relation. 
% However, the LSR fails to predict the relation between $<$ \textit{New Ireland}, \textit{World War II} $>$, which is labeled as \textit{NIL} type. It is explainable as mention \textit{World War II} attached \textit{New Ireland} with a high score, which indicates that these two mentions have certain kind of relations.
However, LSR also attends to the mention \textit{New Ireland} with a high score, thus failing to predict that the entity pair $\langle$\textit{New Ireland}, \textit{World War II}$\rangle$ actually has no relation (\textit{NIL} type). %\footnote{More cases are provided in supplementary materials.}. 

\section{Related Work}
%\vspace{-1mm}
\label{sec:5}

% Our work builds on a rich line of recent efforts on relation extraction models and graph convolutional networks.

\paragraph{Document-level relation extraction.} 
Early efforts focus on predicting relations between entities within a single sentence by modeling interactions in the input sequence \citep{Zeng2014RelationCV, Wang2016RelationCV, Zhou2016AttentionBasedBL, Zhang2017PositionawareAA, Guo2020Forests} or the corresponding dependency tree \citep{Xu2015SemanticRC, Xu2015ClassifyingRV, Liu2015ADN, Miwa2016EndtoEndRE, Zhang2018GraphCO}. These approaches do not consider interactions across mentions and ignore relations expressed across sentence boundaries. Recent work begins to explore cross-sentence extraction \citep{Quirk2017DistantSF, Peng2017CrossSentenceNR, Gupta2018NeuralRE, Song2018NaryRE, song-etal-2019-leveraging}. Instead of using discourse structure understanding techniques \citep{LiuDRS, Lei2017Swim, Lei2018Linguistic}, these approaches leverage the dependency graph to capture inter-sentence interactions, and their scope is still limited to several sentences. More recently, the extraction scope has been expanded to the entire document \citep{Verga2018SimultaneouslyST, Jia2019DocumentLevelNR, Sahu2019IntersentenceRE, christopoulou2019connecting} in the biomedical domain by only considering a few relations among chemicals. Unlike previous work, we focus on document-level relation extraction datasets \citep{yao2019DocRED, Li2016BioCreativeVC, wu2019renet} from different domains with a large number of relations and entities, which require understanding a document and performing multi-hop reasoning. 

\paragraph{Structure-based relational reasoning.}
Structural information has been widely used for relational reasoning in various NLP applications including question answering \citep{dhingra-etal-2018-neural, de2018question, song2018exploring} and relation extraction \citep{Sahu2019IntersentenceRE, christopoulou2019connecting}. \citet{song2018exploring} and \cite{de2018question} leverage co-reference information and set of rules to construct document-level entity graph. GCNs \citep{Kipf2016SemiSupervisedCW} or GRNs \citep{Song2018AGM} are applied to perform reasoning for multi-hop question answering \citep{welbl2018constructing}. \citet{Sahu2019IntersentenceRE} also utilize co-reference links to construct the dependency graph and use labelled edge GCNs \citep{Marcheggiani2017EncodingSW} for document-level relation extraction. Instead of using GNNs, \citet{christopoulou2019connecting} use the edge-oriented model \citep{christopoulou2018walk} for logical inference based on a heterogeneous graph constructed by heuristics. Unlike previous approaches  that use syntactic trees, co-references or heuristics, LSR model treats the document-level structure as a latent variable and induces it in an iteratively refined fashion, allowing the model to dynamically construct the graph for better relational reasoning.

\vspace{-1mm}
\section{Conclusion}
\vspace{-1mm}
We introduce a novel latent structure refinement (LSR) model for better reasoning in the document-level relation extraction task. Unlike previous approaches that rely on syntactic trees, co-references or  heuristics, LSR dynamically learns a document-level structure and makes predictions in an end-to-end fashion. There are multiple avenues for future work. One possible direction is to extend the scope of structure induction for constructions of nodes without relying on an external parser. 
%  and show its superiority on three datasets.
\vspace{-1mm}
\section*{Acknowledgments}
\vspace{-1mm}
We would like to thank the anonymous reviewers for their thoughtful and constructive comments. This research is supported by Ministry of Education, Singapore, under its Academic Research Fund (AcRF) Tier 2 Programme (MOE AcRF Tier 2 Award No: MOE2017-T2-1-156). Any opinions, findings and conclusions or recommendations expressed in this material are those of the authors and do not reflect the views of the Ministry of Education, Singapore.
%We also thank Dekun Wu, Xu Wang, Sicong Leng and Zheng Zhao for discussions.
\bibliography{acl2020}
\bibliographystyle{acl_natbib}

\end{document}